# Cluster-based ensemble learning for wind power modeling with meteorological wind data

Hao Chen*

Department of Technology and Safety, UiT The Arctic University of Norway, Tromsø, Norway

* = corresponding author, hao.chen@uit.no

## Abstract

Optimal implementation and monitoring of wind energy generation hinge on reliable power modeling that is vital for understanding turbine control, farm operational optimization, and grid load balance. Based on the idea of similar wind condition leads to similar wind power; this paper constructs a modeling scheme that orderly integrates three types of ensemble learning algorithms, bagging, boosting, and stacking, and clustering approaches to achieve optimal power modeling. It also investigates applications of different clustering algorithms and methodology for determining cluster numbers in wind power modeling. The results reveal that all ensemble models with clustering exploit the intrinsic information of wind data and thus outperform models without it by approximately 15% on average. The model with the best farthest first clustering is computationally rapid and performs exceptionally well with an improvement of around 30%. The modeling is further boosted by about 5% by introducing stacking that fuses ensembles with varying clusters. The proposed modeling framework thus demonstrates promise by delivering efficient and robust modeling performance.

**Highlights**

- Systematic demonstration of wind power modeling with meteorological data
- Four varying clustering approaches are built and compared in classifying wind data
- Bagging, boosting, and stacking fuse in an orderly and harmonious combination for modeling
- Multiple wind meteorological characteristics are considered in the power modeling

**Word Count:** 7487

**Abbreviations**

| NWP | Numerical Weather Prediction |
|---|---|
| K-means | K-means clusteruing |
| EM | Expectation-Maximization clustering |
| FF | Canopy clustering |
| Canopy | X-means clustering |
| X-means | Farthest First clustering |
| SSE | Sum of Squared Errors |
| BIC | Bayesian Information Criterion |
| NMAE | Normalized Mean Absolute Error |
| NRMSE | Normalized Root Mean Square Error |
| Bagging | Bootstrap aggregating |
| Adaboost | Adaptive boosting |
| REPTREE | Reduced-Error Pruning TREE |
| AdaRF | Adaboost with Random Forest |
| LR | Linear Regression |
| ANN | Artificial three-layer Neural Networks |
| AdaDT | Adaboost Decision Tree |
| CoV | Coefficient of Variation |
| 'Cls-' AdaRF, LR, ANN, AdaDT | Two-layer stacking with four clustering methods as the first and AdaRF, LR, ANN, AdaDT as the second, respectively |
| NCl-AdaRF | Emphasis on AdaRF without a clustering layer (same with AdaRF) |
| FF-AdaRF | Two-layer stacking with FF clustering method and AdaRF |

# 1. Introduction

Wind energy is one of the most commercially viable renewable energy sources, abundant in nature, does not require fossil fuels, and has thus become integral to combating climate change. The Global Wind Energy Council estimates that 355 GW of new capacity will be installed between 2020 and 2024, with almost 71 GW of new installments per year [1]. The rising prevalence of deployed wind energy in the grid brings new challenges for planning electricity generation and dispatching grids because wind power varies intermittently and randomly with prevailing wind conditions.

Establishing an accurate power model for a wind farm, involving among other tasks the empirical mapping of weather data, is related to understanding relationships between wind and its power generation, which in turn is significant to the safe and stable operation of a wind farm and its economic operation [2]. The task is also named farms' non-parametric power curve modeling that can be applied to be a reference profile for the on-line monitoring generation process [3]. More specifically and directly connected this article's aim, wind power modeling is conceptualized as using weather data, more than wind speed, to compute their corresponding wind power.

### 1.1. Related works

Driven by progress in computing affordability and capability and algorithmic advances, wind power can be modeled by physical, statistical, and hybrid methodologies, but there are still potentials to improve these models [4].

Some studies considered meteorological factors in wind power modeling. R. Liu et al. [5] inputted wind speed, wind direction, and air pressure to a power model that was based on multivariable phase space reconstruction, similarity of time-series and linear regression, and demonstrated its superiority for forecasting under conditions where wind power series fluctuate considerably. J. Ma et al. [6] used hourly wind speed and direction at the height of 10 m and 100 m to establish a well-performance model by multivariate empirical dynamic modeling. However, this type of research paid more attention to mapping the relationship between weather data and wind power and lacked weather data explorations and such data's potentials to improve these models.

Ensemble learning also remains a popular approach to improving the modeling since it can reduce the variance and bias of learners. D. Niu et al. [7] established a wind speed-power model with wavelet decomposition and weighted random forest optimized by the niche immune lion algorithm. The model was subsequently tested in two empirical analyses. Y. Dong et al. [8] processed input data with wavelet packet decomposition and applied stacking ensemble by evaluating correlation coefficient between base learners for wind power forecast modeling. The model proudly showed the ensemble edge when the base learners are with good accuracy and low correlations between each other. However, these studies were mainly algorithm-oriented and generally seldom considered wind data's inner characteristics in detail; meanwhile, their usage of decomposition to handle data increased the time complexity in modeling.

Wind has some internal trends that can be understood through data mining approaches. There were some studies on clustering technique applications in wind power modeling. V. Kushwah et al. [9] found that the clusters of time series data showed identical trend components in wind speed data and used cluster-based statistical modeling technique. It showed better performance than other statistical ones. K. Wang et al. [10] clustered Numerical Weather Prediction (NWP) data, consisting of daily wind speed, pressure, humidity, and temperature, by K-means and fed the data into a deep belief network for day-ahead prediction modeling and showed that reduced volatility and sophistication in NWP data made the outperformance. It also revealed the difficulty of tuning hyperparameters in modeling a specific problem. Fortunately, decision tree-type algorithms do not require the adjustment of a large number of parameters. S. Tasnim et al. [11] proposed a K-means cluster-based ensemble regression by linear and support vector regression for wind power forecast modeling, and proved its superiority, up to 17.94% upgrade, by the comparison with no-clustering and several ensemble models in seventy Australian wind sites. The upgrade compared with the baseline is further enlarged to 20.63% by employing a kind of transfer learning approach called multi-source domain adaptation, which includes a weighing method, innovatively calculated with data distributions by K-means clustering, to merge existing sites information for new sites power forecasting [12]. L. Dong et al. [13] utilized cluster analysis of the NWP since wind power and corresponding meteorological data have the characteristic of daily similarity. It suggests that the clustering model is useful in day-ahead modeling of wind power. As evidenced above, the effectiveness of cluster-based wind energy modeling analysis has been validated by multiple relevant models in wind sites worldwide, with engineering applicability and values. Nevertheless, except for the K-means algorithm, other excellent clustering algorithms are rarely employed in this field. Noteworthily, Ref. [13] in this *journal* presented the significance of investigating different clustering approaches in wind power modeling. Wang et al. [14] conducted a self-organizing map clustering for classifying data and used neural networks and support vector machines as base learners to combine a Bayesian model averaging ensembles for analyzing wind power. The model adapts to different meteorological conditions, but its clustering approach and learners are neural networks based and thus have a high temporal complexity. Yet, it remains clear that there is a lack of comparative studies on ensemble learning wind power modeling with different clustering algorithms. Neither is there existing research that combines varying clustering approaches-based stacking ensembles for the modeling tasks. Both of them are addressed in this work.

**1.2. Contribution**

Leveraging the literature review above, a wind farm in the Norwegian Arctic is brought to attention. A wind power modeling framework is proposed, which involves quantifying wind turbulence, clustering meteorological data, and ensemble learning. Firstly, an effective model integrating bagging and boosting is constructed. Secondly, four prominent clustering algorithms are systematically incorporated with models to form layered cluster-based ensembles and the best clustering approach is selected. Finally, stacking is employed to fuse these ensembles with different clusters to establish a more accurate model.

The principal contributions of the paper are thus as follows.

1. This paper experimentally proves the farthest first clustering is a distinctive approach in clustering wind data for power modeling compared to K-means, expectation-maximization, and Canopy clustering algorithms. It is observed that even the worst-performing layered cluster-based ensemble outperforms the one without clustering. This indicates the similarities and dissimilarities in wind data. However, even data are not related to an individual wind turbine; they can still be significantly reflected in wind power in an implicit form.

2. Given diversities in results of different clustering algorithms, it is proposed to fuse layered ensembles with varying clusters with two-layer stacking to formalize a model exceeding the optimal single clustering method. It can efficiently address the complex mapping of incremental nonlinear relationships between meteorological wind data and wind power.
3. A procedure is built for determining the cluster number with a heuristic elbow chart (empirical formula) and an X-means clustering approach. The procedure may be further developed and refined into an improved technique for identifying cluster numbers.
4. Adaboost boosting with random forest bagging as its weak learner is apposite in wind power models. Its power modeling statistically outperforms linear, neural network, and benchmark Adaboost approaches.
5. The quantization of wind turbulence intensities, both wind speed and direction, are applied to wind power modeling in a novel manner. The study finds that both intensities can serve as new features of considering wind volatility in the modeling process.

The remainder of this paper is organized as follows. The wind meteorology and the using data are described in Section 2. In Section 3, an elaborated description of the clustering approaches is presented: how to determine their clusters' number, ensemble learning, the proposed scheme, and statistical ways for comparison. Section 4 shows the experimental procedure applied in the research methodology. Obtained results and discussions are given in Section 5. Finally, Section 6 concludes this work, its implications, and outlooks.

## 2. Wind power meteorology and data preparation

### 2.1. Wind power

Wind power generation is the conversion of wind kinetic energy into electricity. Ignoring losses in the conversion process, the actual output power of wind turbines can be expressed as in (1):

$$P = \begin{cases} 0 & v < v_{min} \\ \frac{1}{2} C_P \rho A v^3 & v_{min} < v < v_n \\ P_n & v_n < v < v_{max} \\ 0 & v > v_{max} \end{cases} \quad (1),$$

where $P$ represents the output power of the wind turbine (W); $C_P$ represents wind energy utilization efficiency; $\rho$ is the air density (kg/m$^2$); $A$ means the effective area swept by the wind turbine blades (m$^2$), $v$ is the wind speed (m/s); $v_{min}$, $v_{max}$, and $v_n$ respectively represent cut-in, cut-off wind speed, and rated wind speed. $P_n$ is the rated wind power for the wind turbine. From (1), the output of a wind turbine is mainly influenced by wind speed, air density, and swept area. Moreover, air density is primarily affected by temperature and pressure [15]. The swept area is related to the wind direction.

## 2.2. Quantification of turbulence in wind

Turbulence arises when airflow flows through uneven landscapes or differences in air density. Turbulence is an immensely complicated flow phenomenon that is highly stochastic and difficult to characterize. In actual wind farm operations, turbulence is generated because of topographic and climate conditions and weak effects between wind turbines: wind is primarily affected by barriers; anomalies in wind occur when wind crosses such places. The impact depends on the height and width of the obstacle. Especially, turbulence has tremendous impacts on wind power production: on similar wind speed conditions, the higher the turbulence intensity, the higher the impact of wind farm output power [16]. The wind turbine's large inertia, including the impeller, whose rotation is behind wind speed change. So, the turbine will not get the theoretical wind force, and the power output goes down. Empirically, at low wind speed, turbulence increases the turbine power production. However, when the wind speed approaches the turbine's furling speed, turbulence reduces the production [17]. Nevertheless, turbulence is rarely considered in machine learning wind energy models. A published article in the *journal* [18] comparatively examined the effects of five popular learning algorithms and nine atmospheric variables on wind turbine power generation and found by statistical tests: first, for these five benchmark algorithms, the selection of atmospheric features for wind power modeling is more crucial; second, the top five features that are most influential for modeling are, in order, wind speed, turbulent kinetic energy, temperature, turbulence intensity, and wind direction. However, turbulent kinetic energy is seldom recorded by wind sites due to its measurement complexity. So, the turbulence intensity is considered as an input feature in this study.

Turbulence intensity, defined as wind speed standard deviation divided by the mean value over a short period [19], is the principal characteristic quantity of wind speed volatility. The turbulence intensity to direction is also applied as a quantitative tool to define turbulence behavior in wind direction. The turbulence intensities are in (2):

$$I_{SP} = \frac{S_{SP}}{SP}, \quad I_D = \frac{S_D}{D} \tag{2}$$

where $I_{SP}$ and $I_D$ are wind turbulence intensity of wind speed and direction; $SP$ is wind speed and $S_{sp}$ is its standard deviation of the previous ten minutes. $D$ is wind direction index and $S_D$ is also its period standard deviation.

## 2.3. Data preparation

The study centers on a 54 MW wind farm designed in northern Norway, located about 500 km inside the Arctic Circle; it stands out as one of the largest wind farms in the Arctic. This farm's terrain features are a small hill, high steep mountains, and fjords, regarded as complex terrain. The wind power station company offers measurement of wind data with 10 mins temporal resolution. We choose the five-dimensional meteorological wind data (wind speed and its variance, wind direction and its variance, temperature) and power data from 0:00 1st January 2017 to 23:50 31st December 2017. Specifically, we calculate the sine values of wind direction and its standard deviation as indicators of wind direction and its fluctuations. Besides, the turbulence intensities of wind speed and sine value of direction are computed as quantitative indices of wind turbulence. In summary, ten-minute resolution wind data, consisting of wind speed and sine direction and their turbulence intensities, temperature, and pressure, are employed to model wind power.

Because scales of variables in the dataset vary widely, it is worth rescaling the original data into new data with similar proportions for each variable. Data standardization can rescale the variable with a mean of zero and a standard deviation of one. It can increase model convergence speed and improve some algorithms' accuracy, especially in distance-based clustering [20].

## 3. Methodology

### 3.1. Chosen clustering approaches

Cluster analysis is an exploratory data mining technique for extracting useful information from high-dimensional datasets. It searches for hidden patterns that may exist in a dataset and is a type of unsupervised learning approach by grouping similar patterns [21]. Unsupervised learning deals with data vectors training sets without labeled values and attempts to find hidden partitions of patterns. Clustering is a classification method for similar objects into different subsets to make the same subset members have identical attributes [22]. This classification requires quantifying the degree of similarity or dissimilarity between observations. The clustering results are strongly dependent on the kind of similarity metric used [23]. The cluster number is typically unknown and needs to be designated according to prior knowledge or determined by some methods. There are several clustering methods proposed. Ref. [24] offered some aspects when choosing a clustering method. The method should be able to effectively and precisely find the suspected cluster types, and it can resist errors in the datasets; besides, it has the availability of computing. This paper selects four clustering approaches, namely, K-means, expectation-maximization, farthest first, and Canopy. The first one is a baseline method, and the other three can be regarded as competitors.

**K-means:** Among clustering algorithms, the K-means algorithm is one of the most classical and popular algorithms. The K-means, proposed in [25], is a robust and versatile clustering algorithm. The target of the K-means is to categorize observations into $k$ clusters. K-means in this study is associated with Euclidean distance. Given a set of $n$ data points $D = \{x_1, \ldots, x_n\}$ in $\mathbb{R}^d$ and an integer $k$, the K-means problem is to determine a set of $k$ centroids $C = \{c_1, \ldots, c_K\}$ in $\mathbb{R}^d$ to minimize the following error function:

$$E(C) = \sum_{\mathbf{x} \in D} \min_{k=1,\ldots,K} \|\mathbf{x} - \mathbf{c}_k\|^2 \qquad (3).$$

It is a combinatorial optimization that equals finding the partition of the *n* instances in *k* clusters whose associated set of mass centers minimizes Eq.(3) [26].

**EM:** Expectation-Maximization (EM) algorithm is proposed by [27]. It provides a simple, easy-to-implement, and efficient tool for learning parameters of a model [28] and is widely used in data clustering for statistical and machine learning programs. It finds the maximum likelihood or maximum posterior of the parameters in a probabilistic modeling process, where the model relies on latent unobservable variables. The EM firstly initializes distribution parameters, then alternates between two steps: the first step is to compute the expectation of variables based on the assumed initial parameters, and the second is maximization, which gives a maximum likelihood estimate of the current parameters through on the expectation values of the latent variables. The two steps repeat iteratively until the desired convergence is realized. When applying it to clustering, the probabilistic model is established on the probability of each data sample to each cluster and distributes samples to the cluster with the biggest possibility. The EM clustering goal is to maximize the overall probability or likelihood of the clusters.

**FF:** The first utilization of Farthest First (FF) traversal is in [29]. It is an effective greedy permutation method in computational geometry. Its underpinning is traversing a sequence of points in space where the initial point is specifically stochastic. The subsequent points are as remote as possible from the prior chosen set of points. FF clustering is the FF traversal application in clustering, which was introduced in [30]. It is an optimized K-means with an analogous procedure selecting the centroids first and assigning the samples to clusters with the maximum distance. Specifically, number *k* of centroids are generated by stochastically choosing a data as the primary cluster centroid and greedily selecting the data as the second centroid when it is FF from the first centroid. The process is conducted henceforth to *k* times. As soon as all the centroids are recognized, FF assigns all the other data to the cluster in which the data have the nearest feature distances. Different from K-means, FF merely requires one traversal to cluster data. All the cluster centers are real data points, not geometric clustering centroids, and their position is fixed in the computation [31]. In most cases, the speed of clustering is considerably accelerated because fewer reassignments and adjustments are involved. The FF traversal is described in **Algorithm 1**.

**Algorithm 1.** Farthest First clustering Algorithm.
1. Farthest first Clustering (*D*: dataset, *k*: cluster number) {
2. select random data as the first point and first centroid;
3. // searching the data sample that is the farthest from the centroid
4. for (I=2,...,*k*) {
5. for (each remaining data sample in *D*) {
6. calculate the total distance to the existing centroids;}
7. select the sample with the largest distance as the new centroid;
8. label the centroids as {$c_1, c_2, ...., c_k$}}
9. //assignment the rest points {$p_1, p_2, ...., p_n$}

10. for (each point $p_i$) {
11. calculate the distance function dist to each fixed cluster centroid;
12. realize min {dist($p_i$, $c_1$), dist ($p_i$, $c_2$), …, dist ($p_i$, $c_k$)]}
13. put it to the cluster with minimum distance;} }

**Canopy:** Canopy clustering was introduced in 2000, and its key idea was using a cheap, approximate distance measure to divide the data into subsets efficiently. This clustering decreases computing time over K-means and EM clustering methods by more than an order of magnitude and reduces error on large datasets [32]. Unlike K-means that only uses one distance, the Canopy algorithm uses two thresholds larger the loose distance $T_1$ and smaller the close distance $T_2$. It begins with removing a random point r sample from the original dataset and starts a *canopy* centered at *r*. Approximate all distances between *r* to remaining data $r_i$. If the distance is less than $T_2$, place $r_i$ in *r* canopy. If the distance is less than $T_1$, remove $r_i$ from a dataset. Repeat these steps until there is no more data to be clustered. However, the Canopy needs tuning the distance parameters and according to [33], $T_1$ and $T_2$ can be obtained approximately using a heuristic based on attribute standard deviation.

## 3.2. Determining the cluster number

While various clustering methods are available, all of the mentioned methods need the cluster number before the clustering procedure. It is necessary to estimate the number due to the resulting partition of the data relays on its specification. This paper combines three methods to find a suitable cluster number for our meteorological wind data.

There is an empirical formula [34] to find the cluster number *k*. It is useful to check the range of *k* since it is not a precise approach.

$$k = 1 + 3.2 \log_{10} n \qquad (4),$$

where *n* is the number of data points. This formula is still inaccurate but can provide a reference value for seeking *k*.

The elbow method is an inelegantly heuristic and visual technique for choosing cluster numbers [35]. The elbow principle's elementary idea is that the total sum of squared errors, the smaller value means the more convergent result, between the sampling point in each cluster and the centroid, are calculated with a series of *k* values. When the setup cluster number approximates the actual cluster number, the sum of squared errors will decrease swiftly. As the setup cluster number continues to grow, the Sum of Squared Errors (SSE) will also continuously descend, but slower [36]. Intuitive observation of turning points from elbow plots is sometimes inaccurate. Still, it can provide a reasonable interval for the value of *k*.

To find the exact value of *k*, the X-means approach offers an effective way.

**X-means:** It is a variation of K-means clustering and can automatically determine the optimal cluster number in a dataset. It refines cluster assignment by repeatedly attempting subdivision segments and keeping the best resulting splits. It searches the space of cluster locations and the cluster number to optimize the Bayesian Information Criterion (BIC) measure [37]. The main parameters for X-means are the lower and upper bounds of the cluster number. It includes two steps that are repeated until they reach the required convergence. Primarily, the K-means algorithm is utilized to cluster the given dataset. Each cluster centroid is divided into two parts in opposite directions along a stochastic vector. The K-means algorithm is locally operated within the old cluster and generates two new clusters. By comparing the BIC scores of the original clustering structure and a new one, the splitting is made or not. The idea is splitting a single cluster into two clusters increases the BIC score, having two clusters is more probable rather than one. When *k* reaches the set upper bound, the splitting stops and the algorithm reports BIC scores with each *k* value.

### 3.3. Modeling ensemble learning algorithm

The fundamental idea behind ensemble learning is to ensemble multiple algorithms or models to achieve an integrated model with better predictive performance [38]. The ensemble method can tactfully partition the dataset into small datasets, train them separately, and then combine them with some strategies. The main strategies can be categorized into three groups: Boosting, Bootstrap aggregating (shortened as Bagging), and Stacking.

In the bagging procedure, new training sets are formed by taking from the original training set with a put-back. The average method for each new result of the training set is applied to get the final result in a regression. Random forest [39] is an efficient bagging algorithm that uses decision trees as its base learner and offers decent performance and low computing costs. It is an improvement in the decision tree algorithm, essentially in which multiple decision trees are merged. The creation of each tree depends on an independent bagging subset. Each tree in the forest has the same probabilistic distribution. The final regression value can be determined by averaging each predictive value from each tree. Since random forest introduces perturbations in sampling and features, it dramatically improves generalization and avoids overfitting. Besides, it can handle high-dimensional data without feature selection, and crucial features are derived during the training process [40].

Boosting [41] is an approach that boosts weak learners to strong learners. Adaptive boosting (shortened as Adaboost) is a representative boosting algorithm. It continually builds weak learners to emphasize (with bigger weights) on samples mislearned in the prior learner until the number of learners reaches the setup value or the loss function reaches a threshold. For the regression problem, the weighted average is used to obtain eventually predicted values. AdaBoost is highly accurate and can adequately construct weak learners and is not susceptible to overfitting. Meanwhile, it is sensitive to anomalous samples (which may receive large weights in iterations), affecting the performance of strong learners. For the numeric output for the strong learner $h_i(\mathbf{x}) \in \mathbb{R}$, weighted averaging (5) is used for the final result [42].

$$H(\mathbf{x}) = \sum_{i=1}^{M} w_i h_i(\mathbf{x})  \tag{5}$$

where $w_i$ is the weight of a weak learner and $w_i \geq 0, \sum_{i=1}^{T} w_i = 1$.

Stacking is a representation learning that can extract valid features from data by employing meta-learning algorithms to learn how to combine predictions from many base learners optimally. Several different base models are first trained with the original dataset. A new model named meta-learner is then trained with each of the previous models' outputs to get a final output [43]. The stacking result is typically better than its single base learned since the fusional ensemble combines varying types of base learners. The applied stacking is shown in **Algorithm 2** [44].

**Algorithm 2.** Stacking Algorithm with four base learners and one meta-learner.

**Input:** Dataset $D = \{(x_1, y_1), (x_2, y_2), ..., (x_m, y_m)\}$

Base learner varying clustering approaches-based Adaboost algorithms $\mathfrak{L}_1, ..., \mathfrak{L}_4$;

Meta-learner linear regression $\mathfrak{L}$

**Process:**
1. **for** $t = 1,2,3,4$ **do**
2.     $h_t = \mathfrak{L}_t(D)$;
3.     // Train base learners by $\mathfrak{L}_t$
4. **end for**
5. // Generate training set for meta-learner
6. $D' = \emptyset$;
7. **for** $i = 1,2,...,m$ **do**
8.     **for** $t = 1,2,3,4$ **do**
9.         $z_{it} = h_t(x_i)$;
10.     **end for**
11.     $D' = D' \cup \left((z_{i1}, z_{i2}, z_{i3}, z_{i4}), y_i\right)$
12. **end for**
13. // Meta-learner $h'$ is established
14. $h' = \mathfrak{L}\left(D'\right)$

**Output:** $H(x) = h'(h_1(x), h_2(x), ..., h_T(x))$

A two-layer assemblage structure, can be categorized as a kind of layered cluster-based or oriented ensemble named by [45], for regression is adopted to optimally incorporate clustering results generated separately by the four above clustering approaches into the Adaboost mechanism. The ensemble structure is with excellent learning ability and prediction accuracy by mapping the first-layer clustering to the second-layer ensemble regression [21].

### 3.4. Proposed modeling approach

For clustering approach comparisons, a proposed framework is displayed in Fig. 1. It is inspired by wind energy meteorology, clustering approaches, and ensemble learning. The framework is a two-layer architecture, with four clustering algorithms in layer 1 and Adaboost in layer 2. Specifically, random forest is the weak learner for the Adaboost and Reduced-Error Pruning TREE (REPTREE) [49] is introduced to replace the decision tree in the random forest to reduce overfitting may be caused by the complicated ensemble model structure.

Take K-means clustering as an example: Layer 1. Using Section 3.2 to identify the cluster number $k$; clustering the wind data into $k$ clusters. Layer 2. Employing each cluster to train the Adaboost to learn Adaboost and establish $k$ submodels with labels. Subsequently, the test data, one by one, are classified into an existing cluster and loaded into the trained Adaboost submodel corresponding to the cluster for wind power modeling, and overall performance is calculated.

Analogously, the above procedure is also applied to EM, FF, and Canopy clustering approaches. More experimental details are presented in Section 4.

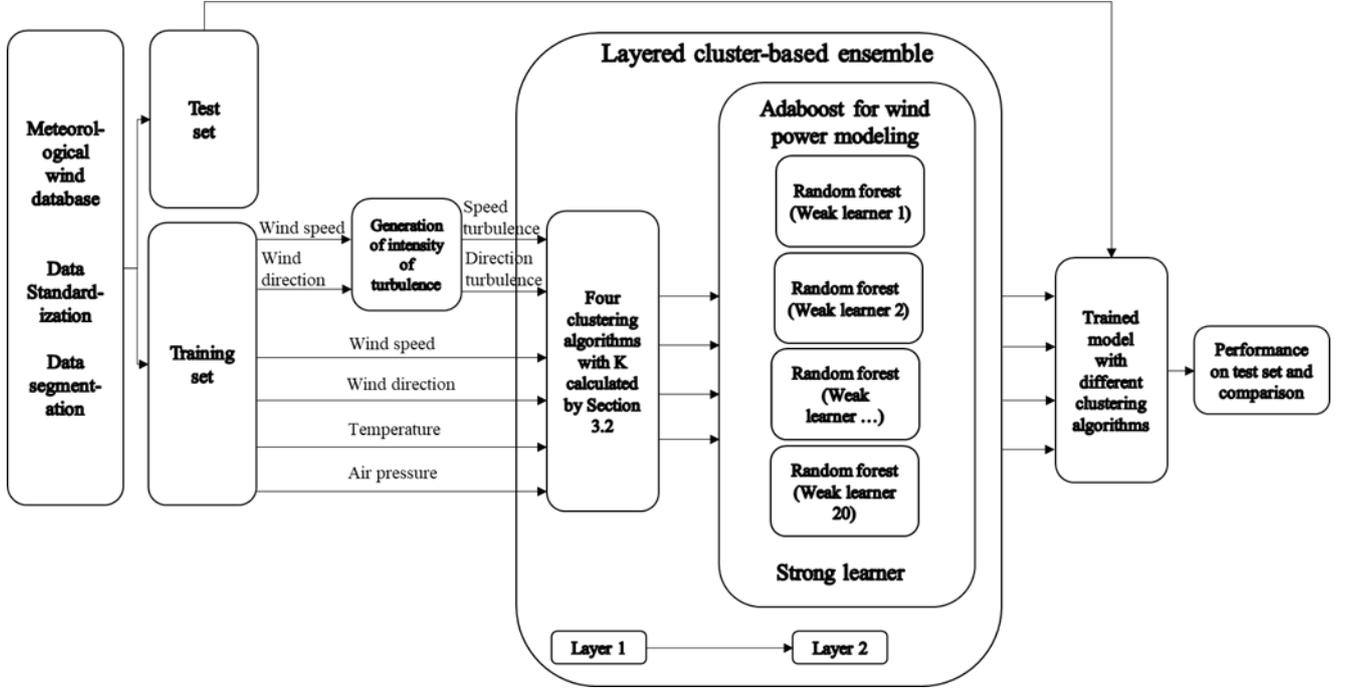

Fig. 1. The procedure of the proposed approach for wind power modeling.

Regarding stacking ensemble modeling, a novel method is put forward. It also consists of two layers, where the first layer is base learners (Adaboost models with the four different clustering algorithms) and the second layer is linear regression Eq.(5) with Tikhonov regularization $\lambda \| w \|_1$ (also named ridge regression [46] Eq.(6) to avoid overfitting caused by the complex model structure [47]). The reasons for this configuration are: 1. The first layer has diversity in the layered ensembles based on four clustering algorithms and may extract data deep features and transmit them to the second layer. 2. The major risk of the second layer is that it learns the generated data from the first layer and is vulnerable to overfitting, so linear regression with a regular term is the learning algorithm in this layer. The first layer procedure is the same as Fig.1. It generates four sets of simulated power on training and test sets. Subsequently, the second layer uses the measured power and four generated power sets as the dependent and independent variables, respectively, to build ridge regression on the training set and employs the learned regression to predict the power with the simulated test power on the test sets.

$$f(x) = w^\top x + b \qquad (6),$$

with a loss function $J = \frac{1}{n}\sum_{i=1}^{n}(f(x_i) - y_i)^2 + \lambda \| w \|_1$:

$$\begin{aligned}\min_{w,b} \quad & \frac{1}{n}\sum_{i=1}^{n}(w^\top x_i + b - y_i)^2 \\ \text{s.t.} \quad & \| w \|_1 \leq t\end{aligned} \qquad (7).$$

## 3.5. Model evaluation metrics and multiple comparisons

Two metrics are utilized in evaluating the performance of different models in the test set. The first one is Normalized Mean Absolute Error (NMAE), the second is Normalized Root Mean Square Error (NRMSE). They are negatively oriented, which means the smaller value is related to better performance. The NRMSE assigns a higher weight to bigger errors because of the square calculation, meaning it punishes substantial prediction errors and points out whether the regression has noticeable error variance.

$$NMAE = \frac{\sum_{i=1}^{n}|prediction_i - observation_i|}{n} / \frac{\sum_{i=1}^{n} observation_i}{n} \tag{8}$$

$$NRMSE = \sqrt{\frac{\sum_{i=1}^{n}(prediction_i - observation_i)^2}{n}} / \frac{\sum_{i=1}^{n} observation_i}{n} \tag{9}$$

Two statistical approaches are used to check whether there are statistically significant differences between the model's performance. The Friedman test is used to check for differences in performance across multiple trials [48]. It tests column effects after adjusting for possible row effects.

$H_o$: The column data do not have a significant difference.

$H_a$: They have a significant difference.

Its statistic $F$ is shown as:

$$F = \frac{12n}{k(k+1)} \left[ \sum_{i=1}^{k} r_i^2 - \frac{k(k+1)^2}{4} \right] \tag{10}$$

where $k$ is the number of columns, $r_i$ is the mean value of row $i$, which follows $\chi^2_{(k-1)}$ under $H_o$.

Besides, the Tukey method is for computing Confidence Intervals between means of two populations, it is expressed:

$$(\bar{Y}_1 - \bar{Y}_2) \pm \frac{q_{k,n-k,1-\alpha}}{\sqrt{2}} \cdot \sqrt{MSE} \cdot \sqrt{\frac{1}{n_1} + \frac{1}{n_2}} \tag{11}$$

where $q$ is the Gaussian $q$-distribution, $k$ is the number of populations, and $n$ is its total size, and $MSE$ donates the Mean Square Error within groups.

## 4. Experiment setup

This study extracts meteorological wind data from the Norwegian Water Resources and Energy Directorate, including a few abnormal negative values, at which the wind farm did not generate electricity but consumed grid power. All weather data are standardized as inputs to the models. First, the wind data are divided into training, accounting for 90%, and test sets with 10%. To fully apply the data, avoid overfitting and improve generalization in modeling [49], 10-fold cross-validation is used in the training set. Then, harness weather data in the test set to calculate the corresponding wind power that are compared to the actual power data to obtain performance metrics.

For the benchmark model, the processed training data are directly fed into the Adaboost with random forest (Layer 2 in Fig. 1) (AdaRF). The number of iterations is set to 100. Random forest is the Adaboost inner weak learners, and the number of REPTREE in each random forest is set to 10. The competitors are linear regression (LR), artificial three-layer neural networks (16 nodes in the hidden layer, which is found by grid search from 6 to 20) (ANN), and Adaboost with 20 decision trees as its weak learners (AdaDT).

Regarding the ensemble model based on clustering approaches, the range of cluster numbers for the weather data is first found by the elbow graph and empirical formula (4). Its exact value is determined using the X-means clustering method. Then, using the four aforementioned clustering approaches to group the data in the training set and categorize the test data into established clusters to find the best performing clustering algorithm. Finally, stacking is employed to combine layered cluster-based ensembles with different clustering algorithms to further explore avenues to upgrade power modeling.

In this study, wind power modeling is realizing the relationship between wind power and wind weather $W_t$. The model is shown in (12).

$$\hat{P}_t = f_t(W_t) + e \qquad (12),$$

in which

$$W = [V, IV_{turbulance}, sin(\theta), Isin_{turbulance}(\theta), T, P] \qquad (13),$$

where $\hat{P}_t$ is modeling wind power; $f_t(.)$ is the model that needs to be implicitly realized; $W_t$ represents weather data that will be clustered by the foregoing four clustering approaches, and $e$ is the model error.

## 5. Experiments and results

### 5.1. Feature ranking and comparison for modeling without clustering

The standardized training wind data are firstly harnessed to establish a multivariate linear wind power regression model to check the feature attributing degree. The diagnosis ($T$ statistic and its corresponding two-tailed $p$-value [50]) for the interpret of each feature is shown in Table. 1.

Table 1. The wind features selected by the statistical diagnosis of linear regression.

| Futures | $V$ | $IV_{turbulance}$ | $Sin(\theta)$ | $Isin_{turbulance(\theta)}$ | $T$ | $P$ |
|---|---|---|---|---|---|---|
| $T$ statistic; $p$-value | 250.79; <0.0001 | 7.84; <0.0001 | -21.77; <0.0001 | 3.02; 0.0025 | 23.68; <0.0001 | 0.67; 0.5040 |

Note: the term is shown '$T$ statistic; $p$-value'. The $H_0$ is the interpret equals zero and its $H_a$ is the term is not zero; when the $p$-value is smaller than the set confidence level 0.05; the $H_0$ is rejected and the feature attributes to the linear model.

All meteorological features are statistically significant in the linear modeling, excluding pressure $P$. The feature's importance may be approximatively ranked by absolute values of $T$ statistics in a descending scale as $V$, $Sin(\theta)$, $T$, $IV_{turbulance}$, $Isin_{turbulance(\theta)}$, $P$. Although pressure does not contribute to the linear regression, all above meteorological features are still accounted for in modeling as pressure values are rather stable and the presented models are clustering and tree models demanding low computations and feature selection.

To enhance the verifiability of modeling results, the year is split into four quarters, Q1, Q2, Q3, Q4, for power modeling individually. The statistical variability among quarterly data is firstly analyzed in Table. 2. Statistics and distribution disparities between meteorological wind and power quarterly datasets can be summarized, and quarterly data differ from the yearly. Therefore, separately modeling on these datasets can strengthen the proposed approach's credibility and the conclusions' reproducibility.

Table 2. The statistics of the yearly and quarterly wind data.

| Statistics / Dataset | Average | Standard deviation | Skewness | Kurtosis |
|---|---|---|---|---|
| **Year** | <0.0001 | 0.9983 | 4.9493 | 100.7258 |
| **Q1** | 0.1988 | 0.8981 | 3.9305 | 87.7615 |
| **Q2** | -0.0283 | 0.8729 | 5.4827 | 129.5278 |
| **Q3** | -0.0356 | 0.8421 | 5.4604 | 118.6208 |
| **Q4** | -0.1350 | 0.8873 | 4.7543 | 113.6719 |
| *CoV for Statistics* | -53584 | 0.0586 | 0.1157 | 0.1316 |

Note: The different variables are standardized to similar scales, so the statistics of the various variables in the dataset are averaged and shown. Coefficient of Variation (CoV) is defined as the ratio of standard deviation to mean.

The four quarterly and yearly training data are separately entered into the proposed AdaRF, benchmarking LR, ANN, and AdaDT to map the relationship between wind data and wind power. Fig.2 shows the results. Both NMAE and NRMSE increase significantly as time grows. The NMAE and NRMSE of AdaRF are significantly lower than the results obtained from multivariate linear regression. The average NMAE and NRMSE are decreased by 52.98% and 46.31%, respectively. The AdaRF decrease in NMAE and NRMSE (corresponding to the model boosting) is also evident when comparing it against ANN (NMAE 19.54% and NRMSE 10.43%) and AdaDT (NMAE 29.31% and NRMSE 17.95%). Fig.3 displays the modeling power of a day, from which AdaRF appears close to real values, but with several errors in points. These mean the proposed AdaRF enables accurate power modeling based on weather data but still leaves scope for refinement.

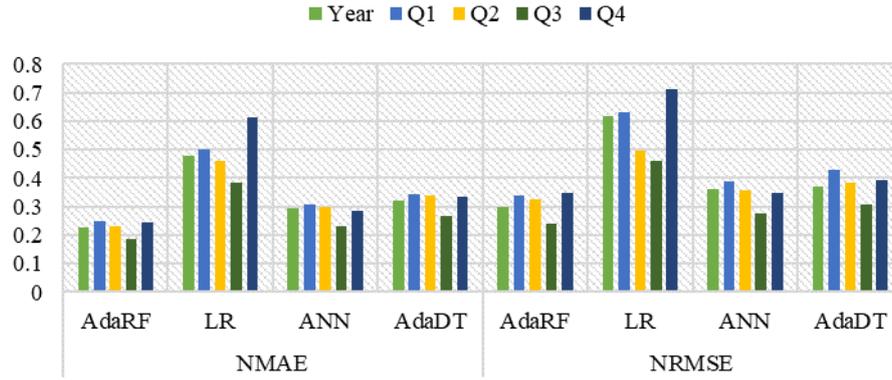

Fig. 2. Comparison for power modeling without clustering.

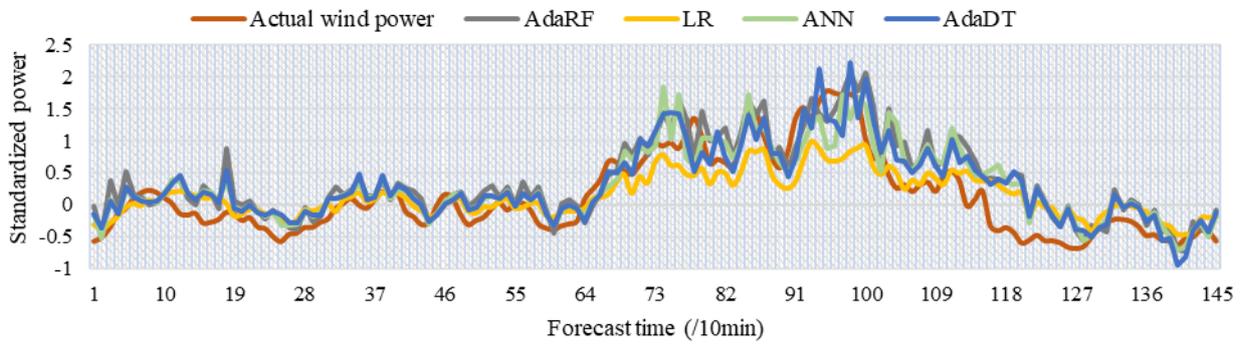

Fig. 3. The wind power curving fitting modeling results in the wind farm.

## 5.2. Determination of appropriate cluster number

The size of the yearly dataset is 52,560; $k$ is calculated to be approximately 16 in (4). Selecting this value as the midpoint, the total Sum of Squared Errors (SSE) of K-means is calculated with a starting point of $k$ equals 2 and an endpoint of $k$ is 30. The elbow plot is drawn and displayed in Fig. 3. It is vague to determine the precise value of elbow point for the total sum of squared errors since the process is by intuition and experience; however, Fig. 3 still shows an interval, the cluster number $k\in[10,20]$, in which the decline of SSE begins to flatten from steep, the elbow point belongs.

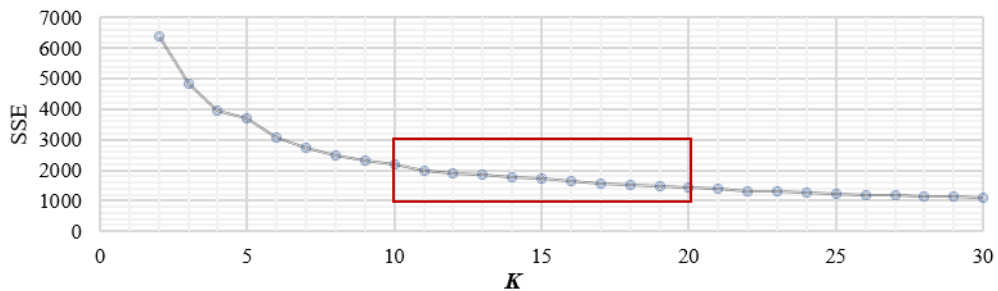

Fig. 4. The elbow plotting for finding cluster number $k$.

To demonstratively find the precise value of *k*, X-means approach is harnessed. The lower and upper bounds of *k* are set as 10 and 20 according to the interval formerly gotten from Fig. 3. The optimized BIC score is 69,956.78 with a proper *k* value for the meteorological wind data equaling 11.

Analogously, the *k* values for the four quarterly wind data Q1, Q2, Q3, Q4 are decided as 14, 8, 9, 12, respectively.

## 5.3. Comparison of different clustering approaches in modeling

For the yearly dataset, the four clustering approaches yield varying numbers of samples per cluster, albeit at the same cluster number. Based on the four clustering methods separately with 11 clusters, four complete layered cluster-based ensembles are developed for clustering comparisons. Firstly, the wind weather data with different clustering approaches in the training and test sets are shown in Fig. 5. Each color represents a cluster, and the vertical axis shows the percentage of each number in the total dataset.

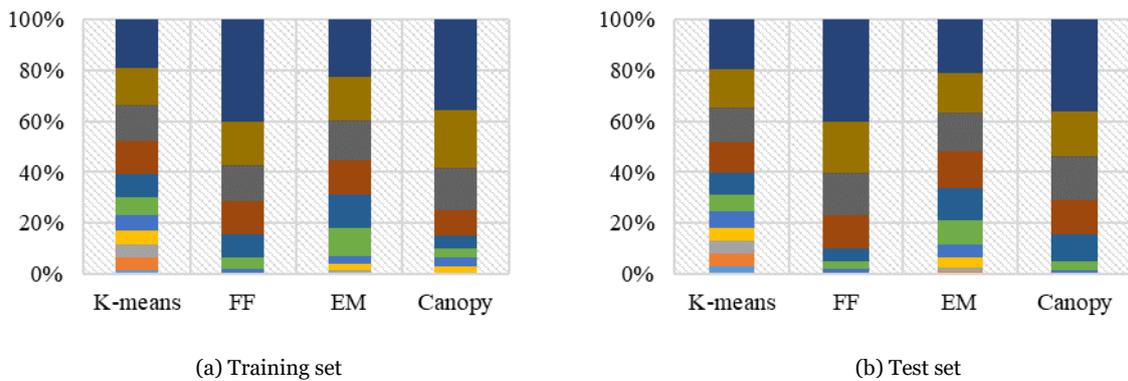

(a) Training set　　　　　　　　　　　　　　(b) Test set

Fig. 5. The clusters number percentage of different clustering approaches.

It is shown that the various clustering methods produce wildly different clustering results, even with the same *k*. The cluster sample number' variance analysis reveals that the K-means method produces more homogeneous clustering than other methods. Even single-digit sample percentages are seen in the FF and Canopy algorithms for the training set. Apart from the K-means, all three other algorithms generate clusters that exceed one-fifth of the sample size. Secondly, the four clustering methods' yearly meteorological wind test data are loaded into the layered cluster-based ensemble for wind power modeling. The NMAE and NRMSE are displayed in Fig. 6.

The model based on FF clustering intuitively presents the smallest NMAE and NRMSE, 32.35% and 33.64% reductions than without clustering; the second smallest being the model with Canopy. The models with these two clustering approaches significantly improve their performance compared to the ones without clustering, while the K-means and EM algorithms also upgrade their models' abilities.

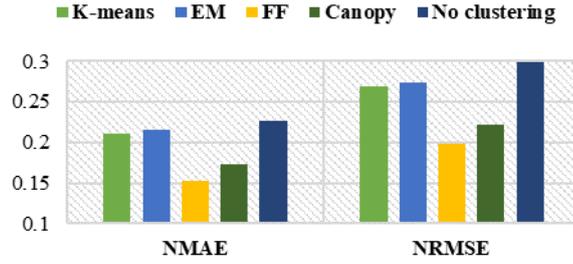

Fig. 6. The NMAE and NRMSE of the power modes with different clustering methods for the yearly dataset.

To further elaborate comparisons of clustering methods, identically, layered cluster-based ensemble modeling with different clustering approaches is conducted on four quarterly wind data, and their NMAE and NRMSE are displayed in Fig. 7. The ranking of the models built on quarterly data is the same as those on the yearly. Strengths in the FF and Canopy clustering algorithms are evident in each quarter. Moreover, a derived result is that the 3rd quarter-power model performs the best, followed by the 2nd quarter. This illustrates a more explicit relationship between wind data and power from April to September, which is consistent with the intuition that the area has milder weather during this period.

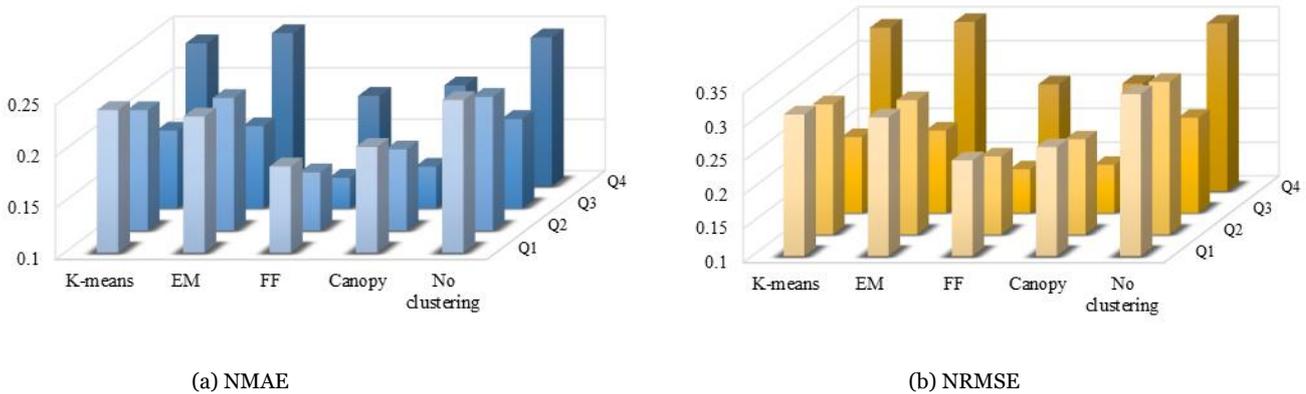

(a) NMAE                                           (b) NRMSE

Fig. 7. The NMAE and NRMSE of the power models with different clustering methods for quarterly data.

Multiple comparisons are conducted between metrics from different cluster-based models between the quarters. The Friedman test *p*-values of NMAE and NRMSE are both 0.0056 and much smaller than the confidence level of 0.05, so the null hypotheses are rejected. It concluded that there are differences between the metrics of the various ensembles.

Table 4 compares the average NMAE and NRMSE for models with different kinds of clustering versus the ones without. Quarterly evenly, the new model reduces NMAE and NRMSE by 13.94% and 17.45%. Furthermore, performance improvement between the two models generally slumps from summer to winter.

Table 4. The performance improvement between average model with clustering and the corresponding one without.

|       | Q1     | Q2     | Q3     | Q4     |
|-------|--------|--------|--------|--------|
| **NMAE**  | 13.66% | 15.06% | 16.29% | 10.74% |
| **NRMSE** | 17.73% | 19.60% | 20.08% | 12.37% |

Regarding the best-performing FF clustering of the yearly dataset, table 5 compares NMAE and NRMSE for the FF-based model to the original model. Both NMAE and NRMSE have about a 23% to 34% reduction, which indicates this clustering approach is twice as good as the average clustering method in our case. Besides, the superiority of FF remains unstable with the quarter: the model boosting results are more noticeable during warm periods compared to those of cold seasons.

Table 5. The performance improvement between the model with FF cluster and the baseline.

|       | Q1     | Q2     | Q3     | Q4     |
|-------|--------|--------|--------|--------|
| **NMAE**  | 25.98% | 31.94% | 30.60% | 23.24% |
| **NRMSE** | 28.83% | 33.85% | 31.68% | 25.88% |

Besides, the Canopy clustering-based approach also displays a reasonably satisfactory result, which improves the modeling performance by average decreasing NMAE and NRMSE over 20% and 25%, respectively. EM and K-means clustering-based approaches have relatively similar performance. Although they are still not as satisfactory as the FF and Canopy, as the above yearly analysis shows.

Collectively, the Tukey method calculates the intervals with 95% confidence of metrics difference; Table 6 shows the bounds of these intervals between the yearly and quarterly models with no clustering and ones with varying clustering algorithms. The upper and lower bounds of the metrics difference between no and FF clustering are positive, indicating that the superiority of the FF is statistically significant across multiple datasets. Moreover, upper bounds of all other differences are greater than lower bounds of absolute values, illustrating that normally distributed differences have positive means, which describes the other cluster-based models are outperforming no clustering in a probabilistic sense. Therefore, the edges of wind data clustering, ranking as FF, Canopy, K-means, EM, before the layered ensemble modeling procedure are demonstrated in our datasets.

Table 6. The bounds for paired comparisons of clustering across yearly and quarterly datasets.

| No clustering |  v.s.        | K-means | EM      | FF     | Canopy  |
|---------------|--------------|---------|---------|--------|---------|
| **NMAE**      | Lower Bound  | -0.0363 | -0.0414 | 0.0176 | 0.0011  |
|               | Upper Bound  | 0.0585  | 0.0534  | 0.1124 | 0.0959  |
| **NRMSE**     | Lower Bound  | -0.0572 | -0.0642 | 0.0122 | -0.0030 |
|               | Upper Bound  | 0.1082  | 0.1012  | 0.1776 | 0.1624  |

### 5.4. Stacking ensemble power modeling

Section 5.1 demonstrates the proposed AdaRF outperforms three other benchmarks (ANN, AdaDT, LR in descending ranking). Section 5.2 illustrates in Fig. 5 the four clustering algorithms yielding highly diverse clusters, and the cluster-based models work better. The AdaRF model is further refined by implementing a two-layer stacking structure (Cls-AdaRF): the first layer takes the four clustering outcomes in Fig. 5 as inputs to AdaRF to generate four layered cluster-based ensembles; the second layer combines these ensembles outputs by linear regression to yield final simulations. Its performance is compared with AdaRF without clustering (NCl-AdaRF) in Section 5.1 and AdaRF with FF clustering (FF-AdaRF) in Section 5.2. The comparison in Fig. 8 shows that Cls-AdaRF decreases more NMAE and NRMSE in percentage than NCl-AdaRF (NMAE 31.89% and NRMSE 34.74%) and FF-AdaRF (NMAE 4.32% and NRMSE 5.71%). Its edge over FF-AdaRF indicates that stacking combined with four various clustering algorithms outperforms the best layered ensemble with a single clustering. These model quarterly variations are similar to those in Section 5.3.

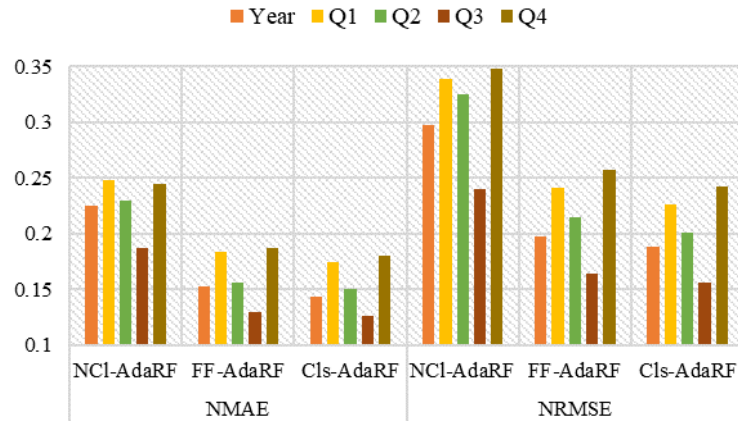

Fig. 8. The NMAE and NRMSE of the power models with different clustering methods for quarterly data.

Likewise, placing benchmark algorithms ANN, AdaDT, and LR into this process eventually yields three cluster-based stacking ensembles, denoted as Cls-LR, Cls-ANN, and Cls-AdaDT. These models, including are run on the datasets separately, and their NMAE and NRMSE are compared to those of Cls-AdaRF. Table 4 shows the difference intervals calculated by Tukey method and reveals that, except for Cls-AdaRF v.s. Cls-ANN in NMAE (the upper bound is considerably close to zero), all the intervals are negative, indicating a 95% statistical significance among datasets for Cls-AdaRF model's strength.

Table 7. The bounds for paired comparisons of stacking across yearly and quarterly datasets.

| Cls-AdaRF | v.s. | Cls-LR | Cls-ANN | Cls-AdaDT |
|---|---|---|---|---|
| **NMAE** | Lower Bound | -0.3735 | -0.1398 | -0.1654 |
| | Upper Bound | -0.2154 | 0.0183 | -0.0073 |
| **NRMSE** | Lower Bound | -0.3896 | -0.1835 | -0.2011 |
| | Upper Bound | -0.2097 | -0.0036 | -0.0212 |

The percentage reductions in metrics for Cls-AdaRF versus other models (corresponding to model boosting) are further calculated and presented in Fig. 9. On average, Cls-AdaRF delivers over 20% improvements over its three competitors (within a standard deviation; Cls-AdaRF outperforms Cls-LR, Cls-ANN, and Cls-AdaDT by about 60%, 25%, and 35%, respectively.).

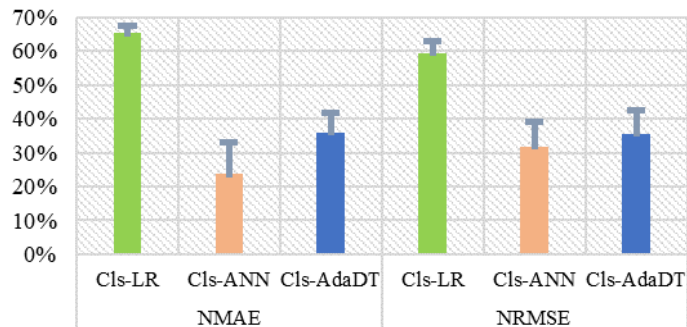

Fig. 9. The NMAE and NRMSE of the power modes with different clustering methods for the yearly dataset.

Altogether, the Cls-AdaRF model is inferred to be a superior wind power model because it does not only outperform AdaRF without clustering but is also better than other stacking models.

## 6. Conclusions

This paper presents an ensemble learning that combines bagging, boosting, and stacking for modeling wind power with meteorological data. To mine the inherent characteristics of the data, four clustering approaches are used to process inputs for the layered ensembles. Then, the layered cluster-based ensembles are fused within the stacking framework. The proposed models' superiority is verified by diverse comparisons.

The AdaRF can accurately implement wind power modeling. This algorithm circumvents issues of an equal weighting of each tree in RF and Adaboost hypersensitivity to outliers, allows each learner to boost incrementally, and eventually creates a model with decent generalizations. The overall performance of the proposed method is on average 33.94% in NMAE and 24.90% in NRMSE lower compared to the benchmarks in the cases excluding clustering.

As no standard methods for identifying the cluster number exist, this paper uses a heuristic elbow graph, empirical formula, and X-means clustering algorithm to preciously determine the implied number for meteorological data. Especially the number for the yearly dataset is 11, which is close to the month's number. This result hints that there may be analogous phenomena to measured wind data with monthly periodicity.

A comparative study of AdaRF based on different clustering methods reveals, firstly, that the model with clusters significantly performs better than the model without, regardless of the what clustering approach is employed. This suggests that similarities within the wind power data can correspond to similarities within the weather data. Secondly, among these clustering methods, the model with the FF clustering provides the best modeling results. The reason is that FF is built on finding the data point furthest from the previous centroid as the new one; in other words, emphasizing large differences between clusters. Upon this clustering, the fluctuations among the original meteorological data are considerably diminished, which in turn correspond to a smoother wind power output and increase the accuracy of the wind power model. Considering the fast computability and accuracy of FF, it is suggested that this clustering technique can be applied to ultra-short-term wind power models. Thirdly, Canopy is the speediest among the four clustering methods and also achieves comparative results. Therefore, Canopy can also serve as a favorable clustering approach when wind weather datasets are considerably large.

Finally, the wind power model is further strengthened by using stacking Cls-AdaRF to fuse the layered ensembles with four clustering approaches. It can be interpreted as stacking is representation learning, i.e., effective features are automatically collected from raw data, and fed into the second layer via multiple learners in the first layer. The second layer compiles and aggregates these features through linear regression with a regular term and effectively outputs simulations.

The further research, extrapolated from the above conclusions, is to deeply optimize the base learners of stacking and their combination algorithms to deliver faster and more accurate modeling. The other direction is to incorporate this article's in-the-now power modeling approach with meteorological data with historical wind power to achieve efficacious short-term power forecasting.

## Data Availability

Data for this study are available from the corresponding author provided reasonable requirements are present.

## Acknowledgments

The author thanks Dr. Yngve Birkelund for organizing data and supervision, Dr. Xiuhua Zhu's help in the author's visiting Max Planck Institute for Meteorology, during which the article was revised, and Dr. Bjørn-Morten Batalden, Dr. Abbas Barabadi, and Gunn-Helene Turi for their pertinent comments and valuable support. The publication charge has been funded by a grant from the publication fund of UiT The Arctic University of Norway.

## References

[1]   J. Lee et al., "Global wind report 2019," *Brussels: Global Wind Energy Council (GWEC),* 2020.

[2]   Z. Tian, Y. Ren, and G. Wang, "Short-term wind power prediction based on empirical mode decomposition and improved extreme learning machine," *Journal of Electrical Engineering & Technology,* vol. 13, no. 5, pp. 1841-1851, 2018.

[3]   A. Marvuglia and A. Messineo, "Monitoring of wind farms' power curves using machine learning techniques," *Applied Energy,* vol. 98, pp. 574-583, 2012.

[4]   A. M. Foley, P. G. Leahy, A. Marvuglia, and E. J. McKeogh, "Current methods and advances in forecasting of wind power generation," *Renewable energy,* vol. 37, no. 1, pp. 1-8, 2012.

[5]   R. Liu, M. Peng, and X. Xiao, "Ultra-short-term wind power prediction based on multivariate phase space reconstruction and multivariate linear regression," *Energies,* vol. 11, no. 10, p. 2763, 2018.

[6]   J. Ma, M. Yang, X. Han, and Z. Li, "Ultra-short-term wind generation forecast based on multivariate empirical dynamic modeling," *IEEE Transactions on Industry Applications,* vol. 54, no. 2, pp. 1029-1038, 2017.

[7]   D. Niu, D. Pu, and S. Dai, "Ultra-short-term wind-power forecasting based on the weighted random forest optimized by the niche immune lion algorithm," *Energies,* vol. 11, no. 5, p. 1098, 2018.

[8]   Y. Dong, H. Zhang, C. Wang, and X. Zhou, "Wind power forecasting based on stacking ensemble model, decomposition and intelligent optimization algorithm," *Neurocomputing,* vol. 462, pp. 169-184, 2021.

[9]   V. Kushwah, R. Wadhvani, and A. K. Kushwah, "Trend-based time series data clustering for wind speed forecasting," *Wind Engineering,* p. 0309524X20941180, 2020.

[10]  K. Wang, X. Qi, H. Liu, and J. Song, "Deep belief network based k-means cluster approach for short-term wind power forecasting," *Energy,* vol. 165, pp. 840-852, 2018.

[11]  S. Tasnim, A. Rahman, A. M. T. Oo, and M. E. Haque, "Wind power prediction using cluster based ensemble regression," *International Journal of Computational Intelligence and Applications,* vol. 16, no. 04, p. 1750026, 2017.

[12]  S. Tasnim, A. Rahman, A. M. T. Oo, and M. E. Haque, "Wind power prediction in new stations based on knowledge of existing Stations: A cluster based multi source domain adaptation approach," *Knowledge-Based Systems,* vol. 145, pp. 15-24, 2018.

[13] L. Dong, L. Wang, S. F. Khahro, S. Gao, and X. Liao, "Wind power day-ahead prediction with cluster analysis of NWP," *Renewable and Sustainable Energy Reviews,* vol. 60, pp. 1206-1212, 2016.

[14] G. Wang, R. Jia, J. Liu, and H. Zhang, "A hybrid wind power forecasting approach based on Bayesian model averaging and ensemble learning," *Renewable Energy,* vol. 145, pp. 2426-2434, 2020.

[15] C. Jung and D. Schindler, "The role of air density in wind energy assessment–A case study from Germany," *Energy,* vol. 171, pp. 385-392, 2019.

[16] K. Kaiser, W. Langreder, H. Hohlen, and J. Højstrup, "Turbulence correction for power curves," in *Wind Energy*: Springer, 2007, pp. 159-162.

[17] W. D. Lubitz, "Impact of ambient turbulence on performance of a small wind turbine," *Renewable Energy,* vol. 61, pp. 69-73, 2014.

[18] M. Optis and J. Perr-Sauer, "The importance of atmospheric turbulence and stability in machine-learning models of wind farm power production," *Renewable and Sustainable Energy Reviews,* vol. 112, pp. 27-41, 2019.

[19] M. Türk and S. Emeis, "The dependence of offshore turbulence intensity on wind speed," *Journal of Wind Engineering and Industrial Aerodynamics,* vol. 98, no. 8-9, pp. 466-471, 2010.

[20] A. Ng, "Advice for applying machine learning," in *Machine learning*, 2011.

[21] B. Verma and A. Rahman, "Cluster-oriented ensemble classifier: Impact of multicluster characterization on ensemble classifier learning," *IEEE Transactions on knowledge and data engineering,* vol. 24, no. 4, pp. 605-618, 2011.

[22] P. Berkhin, "A survey of clustering data mining techniques," in *Grouping multidimensional data*: Springer, 2006, pp. 25-71.

[23] A. Serra and R. Tagliaferri, "Unsupervised learning: clustering," *Encyclopedia of Bioinformatics and Computational Biology; Elsevier: Amsterdam, The Netherlands,* pp. 350-357, 2019.

[24] G. W. Milligan, "Clustering validation: results and implications for applied analyses," in *Clustering and classification*: World Scientific, 1996, pp. 341-375.

[25] J. MacQueen, "Some methods for classification and analysis of multivariate observations," in *Proceedings of the fifth Berkeley symposium on mathematical statistics and probability*, 1967, vol. 1, no. 14: Oakland, CA, USA, pp. 281-297.

[26] M. Capó, A. Pérez, and J. A. Lozano, "An efficient approximation to the K-means clustering for massive data," *Knowledge-Based Systems,* vol. 117, pp. 56-69, 2017.

[27] A. P. Dempster, N. M. Laird, and D. B. Rubin, "Maximum likelihood from incomplete data via the EM algorithm," *Journal of the Royal Statistical Society: Series B (Methodological),* vol. 39, no. 1, pp. 1-22, 1977.

[28] C. B. Do and S. Batzoglou, "What is the expectation maximization algorithm?," *Nature biotechnology,* vol. 26, no. 8, pp. 897-899, 2008.

[29] D. J. Rosenkrantz, R. E. Stearns, and I. Lewis, Philip M, "An analysis of several heuristics for the traveling salesman problem," *SIAM journal on computing,* vol. 6, no. 3, pp. 563-581, 1977.

[30] D. S. Hochbaum and D. B. Shmoys, "A best possible heuristic for the k-center problem," *Mathematics of operations research,* vol. 10, no. 2, pp. 180-184, 1985.

[31] R. D. H. Devi, A. Bai, and N. Nagarajan, "A novel hybrid approach for diagnosing diabetes mellitus using farthest first and support vector machine algorithms," *Obesity Medicine,* vol. 17, p. 100152, 2020.

[32] A. McCallum, K. Nigam, and L. H. Ungar, "Efficient clustering of high-dimensional data sets with application to reference matching," in *Proceedings of the sixth ACM SIGKDD international conference on Knowledge discovery and data mining*, 2000, pp. 169-178.

[33] R. R. Bouckaert *et al.*, "WEKA manual for version 3-9-1," *University of Waikato, Hamilton, New Zealand,* 2016.

[34] J. E. Doran, J. Doran, F. E. Hodson, and F. R. Hodson, *Mathematics and computers in archaeology*. Harvard University Press, 1975.

[35] A. Ng, "Clustering with the k-means algorithm," *Machine Learning,* 2012.

[36]	C. Yuan and H. Yang, "Research on K-value selection method of K-means clustering algorithm," *J—Multidisciplinary Scientific Journal,* vol. 2, no. 2, pp. 226-235, 2019.
[37]	 D. Pelleg and A. W. Moore, "X-means: Extending k-means with efficient estimation of the number of clusters," in *Icml*, 2000, vol. 1, pp. 727-734.
[38]	L. Rokach, "Ensemble-based classifiers," *Artificial intelligence review,* vol. 33, no. 1-2, pp. 1-39, 2010.
[39]	L. Breiman, "Random forests," *Machine learning,* vol. 45, no. 1, pp. 5-32, 2001.
[40]	M. Fernández-Delgado, E. Cernadas, S. Barro, and D. Amorim, "Do we need hundreds of classifiers to solve real world classification problems?," *The journal of machine learning research,* vol. 15, no. 1, pp. 3133-3181, 2014.
[41]	L. Breiman, "Bias, variance, and arcing classifiers," Tech. Rep. 460, Statistics Department, University of California, Berkeley …, 1996.
[42]	 A. Bertoni, P. Campadelli, and M. Parodi, "A boosting algorithm for regression," in *International conference on artificial neural networks*, 1997: Springer, pp. 343-348.
[43]	L. Breiman, "Stacked regressions," *Machine learning,* vol. 24, no. 1, pp. 49-64, 1996.
[44]	Z.-H. Zhou, "Ensemble learning," in *Machine Learning*: Springer, 2021, pp. 181-210.
[45]	A. Rahman and B. Verma, "Novel layered clustering-based approach for generating ensemble of classifiers," *IEEE Transactions on Neural Networks,* vol. 22, no. 5, pp. 781-792, 2011.
[46]	G. C. McDonald, "Ridge regression," *Wiley Interdisciplinary Reviews: Computational Statistics,* vol. 1, no. 1, pp. 93-100, 2009.
[47]	S. Džeroski and B. Ženko, "Is combining classifiers with stacking better than selecting the best one?," *Machine learning,* vol. 54, no. 3, pp. 255-273, 2004.
[48]	J. D. Gibbons and S. Chakraborti, *Nonparametric Statistical Inference: Revised and Expanded*. CRC press, 2014.
[49]	 S. Yadav and S. Shukla, "Analysis of k-fold cross-validation over hold-out validation on colossal datasets for quality classification," in *2016 IEEE 6th International conference on advanced computing (IACC)*, 2016: IEEE, pp. 78-83.
[50]	G. A. Seber and A. J. Lee, *Linear regression analysis*. John Wiley & Sons, 2012.